\documentclass{article}
\usepackage[T1]{fontenc} 
\usepackage[utf8]{inputenc} 
\usepackage{ismir,amsmath,cite,url}
\usepackage{graphicx}
\usepackage{color}
\usepackage{subcaption}
\usepackage{paralist}
\usepackage{multicol}
\usepackage{microtype}
\usepackage{amsfonts}
\usepackage{multirow}
\usepackage{booktabs}
\usepackage{tabularx}

\title{Learning To Traverse Latent Spaces for\\ Musical Score Inpainting}

\multauthor
{Ashis Pati$^1$ \hspace{1cm} Alexander Lerch$^1$ \hspace{1cm} Gaëtan Hadjeres$^2$} {
    $^1$ Center for Music Technology, Georgia Institute of Technology, Atlanta, USA\\
    $^2$ Sony CSL, Paris, France\\
{\tt\small ashis.pati@gatech.edu, alexander.lerch@gatech.edu, gaetan.hadjeres@sony.com}}
\def\authorname{Ashis Pati, Alexander Lerch, Gaëtan Hadjeres}

\begin{document}
\maketitle
\begin{abstract}
    Music Inpainting is the task of filling in missing or lost information in a piece of music. We investigate this task from an interactive music creation perspective. To this end, a novel deep learning-based approach for musical score inpainting is proposed. The designed model takes both past and future musical context into account and is capable of suggesting ways to connect them in a musically meaningful manner. To achieve this, we leverage the representational power of the latent space of a Variational Auto-Encoder and train a Recurrent Neural Network which learns to traverse this latent space conditioned on the past and future musical contexts. Consequently, the designed model is capable of generating several measures of music to connect two musical excerpts. 
    The capabilities and performance of the model are showcased by comparison with competitive baselines using several objective and subjective evaluation methods. The results show that the model generates meaningful inpaintings and can be used in interactive music creation applications. Overall, the method demonstrates the merit of learning complex trajectories in the latent spaces of deep generative models. 
\end{abstract}

\section{Introduction} \label{sec:intro}
    Over the last decade, machine learning techniques have emerged as the tool of choice for the design of symbolic music generation models~\cite{fiebrink2016machine} with deep learning being the most widely used~\cite{Briot2018}. Deep generative models have been successfully applied to several different music generation tasks, e.g., monophonic music generation \cite{pmlr-v80-roberts18a,colombo2016algorithmic,sturm2016music}, polyphonic music generation~\cite{yang2017midinet,boulanger2012modeling} and creating musical renditions with expressive timing and dynamics~\cite{oore2018time,huang2019musictransformer}. However, most of these models assume sequential generation of music, i.e, the generated music depends only on the music that has preceded it. In other words, the models rely only on the past musical context. This approach does not align with typical human compositional practices which are often iterative and non-sequential in nature. In addition, the sequential generation paradigm places severe limitations on the degree of interactivity allowed by these models~\cite{Hadjeres2018ARNN, briot2017deep}. Once generated, there is no way to tweak specific parts of the generation so as to conform to users' aesthetic sensibilities or compositional requirements.

    \begin{figure}[t]
        \centerline{
        \includegraphics[width=\columnwidth]{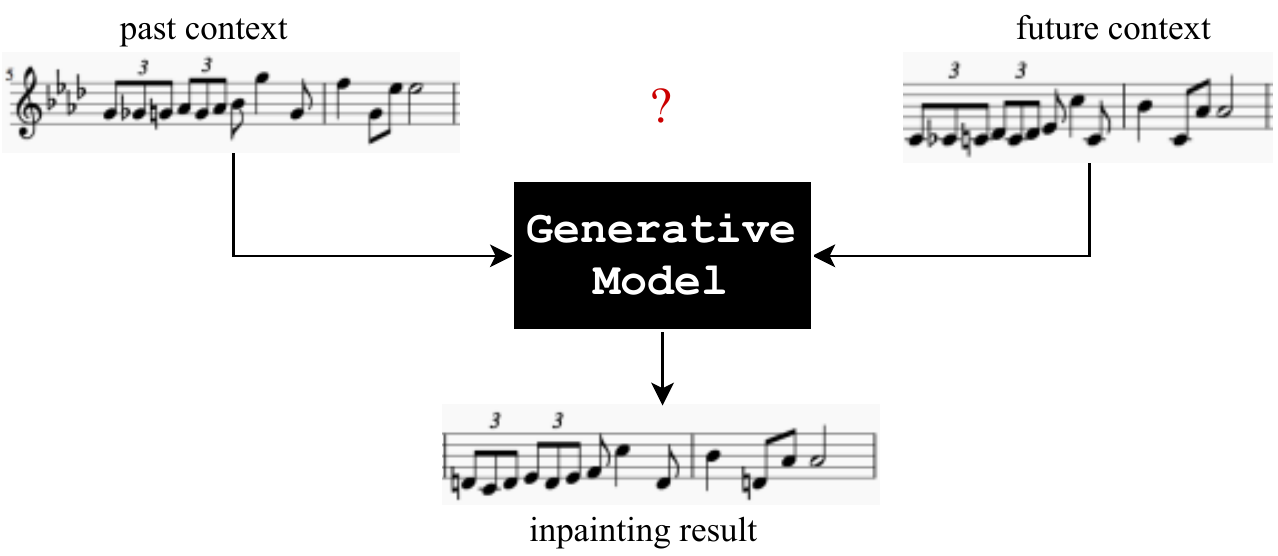}}
        \caption{\textit{Musical Score Inpainting} task schematic. A generative model needs to take past and future musical contexts into account to generate a sequence that can connect them in a musically meaningful manner.}
    \label{fig:inpaint_schematic}
    \end{figure}

    In this paper, we seek to address these problems by incorporating future musical context into the generation process. Specifically, the task is to train models to fill in missing information in musical scores, duly taking into account the complete musical context~---~both past and future. In essence, this is similar to \textit{inpainting} where the objective is to reconstruct missing or degraded parts of any kind of media~\cite{bertalmio2000image}. For music, inpainting has been traditionally used for restoration purposes~\cite{adler2012audio} or to remove unwanted artifacts such as clipping~\cite{bilen2015audio,laguna2016efficient} and packet loss~\cite{perraudin2018inpainting}. However, we investigate models for \textit{Musical Score Inpainting} (see \figref{fig:inpaint_schematic} and \secref{problem_statement}) as tools for music creation which can aid people in
    \begin{inparaenum}[(i)]
        \item getting new musical ideas based on specific styles, 
        \item joining different musical sections together, and
        \item modifying or extending solos. 
    \end{inparaenum} 
    In addition, such models can allow interactive music generation by enabling users to change the musical context and get new suggestions based on the updated context.

    Our main technical contribution is a novel approach for musical score inpainting which relies on \textit{latent} representation-based deep generative models. These models are trained to compress information from high-dimensional spaces, e.g., the space of all $1$-bar melodies, to low-dimensional \textit{latent} spaces. While these latent spaces have been shown to be able to encode hidden attributes of musical data (see \secref{latentspacesmusic}), the primary form of interaction with latent spaces has been using simple operations such as attribute vectors \cite{mikolov2013distributed,carter2017using} or linear interpolations \cite{roberts2018learning,hadjeres2017glsr}. Using the proposed method (see \secref{sec:method}), we demonstrate that Recurrent Neural Networks (RNNs) can be trained using latent embeddings to learn complex trajectories in the latent space. This, in turn, is used to predict how to fill in missing measures in a piece of symbolic music. Our secondary contributions are:
    \begin{inparaenum}[(i)]
        \item a stochastic training scheme which helps model training and generalization (see \secref{train_meth}), and
        \item a novel data encoding scheme using uneven tick durations that allows encoding triplets without substantial increase in sequence length (see \secref{datarep}).
    \end{inparaenum}
    The effectiveness of the proposed method is demonstrated using several objective and subjective evaluation methods in \secref{experiments}.

\section{Related Work} \label{relwork}
    
\subsection{Audio \& Music Inpainting} \label{inpainting}
    The first applications of audio inpainting methods were restoration-oriented~\cite{adler2012audio,bilen2015audio,bilen2015joint,perraudin2018inpainting,toumi2018sparse} using different methods such as matrix factorization~\cite{bilen2015audio}, non-local similarity measures \cite{toumi2018sparse} and audio similarity graphs~\cite{perraudin2018inpainting}. While these techniques have been useful for audio-based tasks, they are not easily extendable to symbolic music.

    For inpainting in the symbolic domain, the early attempts were based on Markov Chain Monte Carlo (MCMC) methods which allowed users to specify certain constraints, e.g., which notes to generate and which to retain~\cite{sakellariou2015maximum,hadjeres2017deepbach}. Another approach, proposed by Lattner et al., used iterative gradient descent to force the output of a deep generative model to conform to a specified structural plan~\cite{lattner2018imposing}. However, methods based on MCMC (which rely on repeated sampling), and those using iterative gradient descent are slow during inference time and hence unsuitable for interactive applications. More recently, Hadjeres et al.~proposed the AnticipationRNN framework~\cite{Hadjeres2018ARNN} which used a pair of stacked RNNs to enforce user-defined constraints during inference. This allowed selective regeneration of specific parts of the music (generated or otherwise) using only two forward passes through the RNN-pair and enabled real-time generations.

\subsection{Variational Auto-Encoders} \label{vae}
    The Variational Auto-Encoder (VAE)~\cite{kingma2013vae} is a type of generative model which uses an auto-encoding~\cite{vincent2008extracting} framework; during training, the model is forced to reconstruct its input. The architecture comprises an encoder and a decoder. The encoder learns to map real data-points $\mathbf{x}$ from a high-dimensional data-space $X$ to points in a low-dimensional space $Z$ which is referred to as the \textit{latent} space. The decoder learns to map the latent vectors back to the data-space. VAEs treat the latent vector as a random variable and model the generative process as a sequence of sampling operations: $\mathbf{z} \sim p(\mathbf{z})$, and $\mathbf{x}\sim p(\mathbf{x}|\mathbf{z})$, where $p(\mathbf{z})$ is a prior distribution over the latent space, and $p(\mathbf{x}|\mathbf{z})$ is the conditional pdf. Variational inference \cite{blei2017variational} is used to approximate the posterior by minimizing the KL-divergence~\cite{kullback1951information} between the approximate posterior $q(\mathbf{z}|\mathbf{x})$ and the true posterior $p(\mathbf{z}|\mathbf{x})$ by maximizing the evidence lower bound (ELBO)~\cite{kingma2013vae}. The training ensures that the reconstruction accuracy is maximized and realistic samples are generated when latent vectors are sampled using the prior $p(\mathbf{z})$.

\subsection{Leveraging Latent Spaces for Music Generation} \label{latentspacesmusic}
    Latent representation-based models such as VAEs have been found to be quite useful for several music generation tasks. Bretan et al.~used the latent representation of an auto-encoder-based model to generate musical phrases~\cite{bretan2016unit}. Lattner et al.~forced the latent space of a gated auto-encoder to learn pitch interval-based representations which improved the performance of predictive models of music~\cite{lattner2018predictive, artz2018audiotoscore}. Latent spaces of music generation models have also been used to explicitly encode and control musical attributes~\cite{hadjeres2017glsr, engel2017latent, pmlr-v80-roberts18a, pati_latent_2019}, inter-track dependencies~\cite{simon2018learning} and musical genre~\cite{brunner2018midivae}. These studies show that trained latent spaces are able to encode hidden attributes of musical data which can be leveraged for different music generation tasks. However, latent space traversals have been relying on simpler methods such as attribute vectors~\cite{mikolov2013distributed,carter2017using} or linear interpolations \cite{roberts2018learning,hadjeres2017glsr}.

\section{Method} \label{sec:method}

\subsection{Problem Statement} \label{problem_statement}
    We define the score inpainting problem as follows: given a past musical context $\mathcal{C}_\mathrm{p}$ and a future musical context $\mathcal{C}_\mathrm{f}$, the modeling task is to generate an inpainted sequence $\mathcal{C}_\mathrm{i}$ which can connect $\mathcal{C}_\mathrm{p}$ and $\mathcal{C}_\mathrm{f}$ in a musically meaningful manner. In other words, the model should be trained to maximize the likelihood $p(\mathcal{C}_\mathrm{i}~|~\mathcal{C}_\mathrm{p},\mathcal{C}_\mathrm{f})$. Without much loss of generality, we assume that $\mathcal{C}_\mathrm{p}$, $\mathcal{C}_\mathrm{f}$, and $\mathcal{C}_\mathrm{i}$ comprise of $n_\mathrm{p}$, $n_\mathrm{f}$, and $n_\mathrm{i}$ measures of music, respectively. 

    \begin{figure}[t]
        \centerline{
        \includegraphics[width=\columnwidth]{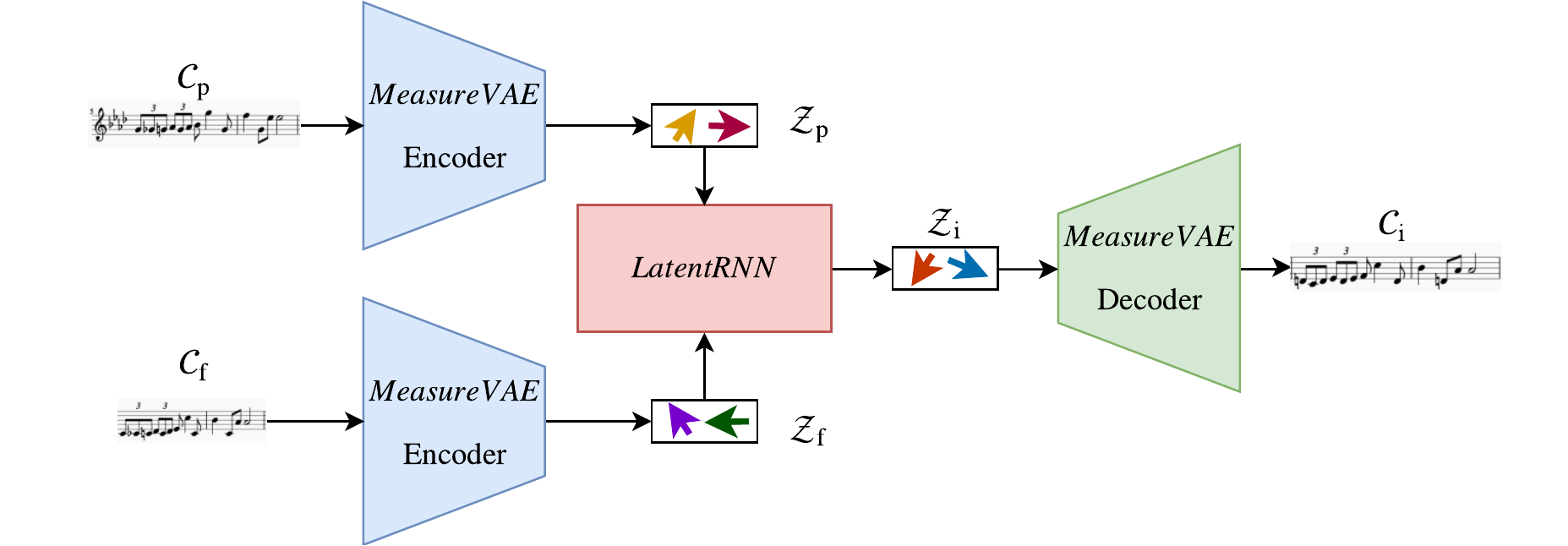}}
        \caption{Schematic of the proposed approach. The pre-trained MeasureVAE encoder is used to convert the past and future context sequences ($\mathcal{C}_\mathrm{p}$ and $\mathcal{C}_\mathrm{f}$) into their respective latent vector sequences ($\mathcal{Z}_\mathrm{p}$ and $\mathcal{Z}_\mathrm{f}$). The LatentRNN learns to traverse the latent space of MeasureVAE to output a latent vector sequence $\mathcal{Z}_\mathrm{i}$ which is passed through the pre-trained decoder to output the inpainted musical sequence $\mathcal{C}_\mathrm{i}$.}
    \label{fig:approach_schematic}
    \end{figure}

\subsection{Approach} \label{approach}
    The key motivation behind the proposed method is that the latent embeddings of deep generative models of music encode hidden attributes of music which can be leveraged to perform inpainting. Firstly, we train a VAE-model, referred to as \textit{MeasureVAE}, to reconstruct single measures of music, i.e., the latent vectors of this model $\mathbf{z} \in Z$ map to individual measures of music. Once trained, the encoder of this model can be used to process sequences $\mathcal{C}_\mathrm{p}$ and $\mathcal{C}_\mathrm{f}$ and output corresponding latent vector sequences $\mathcal{Z}_\mathrm{p}$ and $\mathcal{Z}_\mathrm{f}$. Secondly, we train an RNN-based model, referred to as \textit{LatentRNN}, to take as input the past and future latent vector sequences ($\mathcal{Z}_p$ and $\mathcal{Z}_\mathrm{f}$) and output a third latent vector sequence $\mathcal{Z}_\mathrm{i}$ which can be passed through the decoder of MeasureVAE to obtain $\mathcal{C}_\mathrm{i}$.
    
    Effectively, the LatentRNN model learns to traverse the latent space of the MeasureVAE model so as to connect the provided contexts in a musically meaningful manner. The inference is fast since it only requires forward passes through the two models. This overall approach is shown in \figref{fig:approach_schematic}. We call this joint architecture \textit{InpaintNet}. While we restrict ourselves to $4/4$ monophonic melodic sequences in this paper, the approach can be extended to other time signatures and polyphonic sequences as well. The individual model architectures are discussed next.
 
    \begin{figure}[t]
        \centerline{
        \includegraphics[width=\columnwidth]{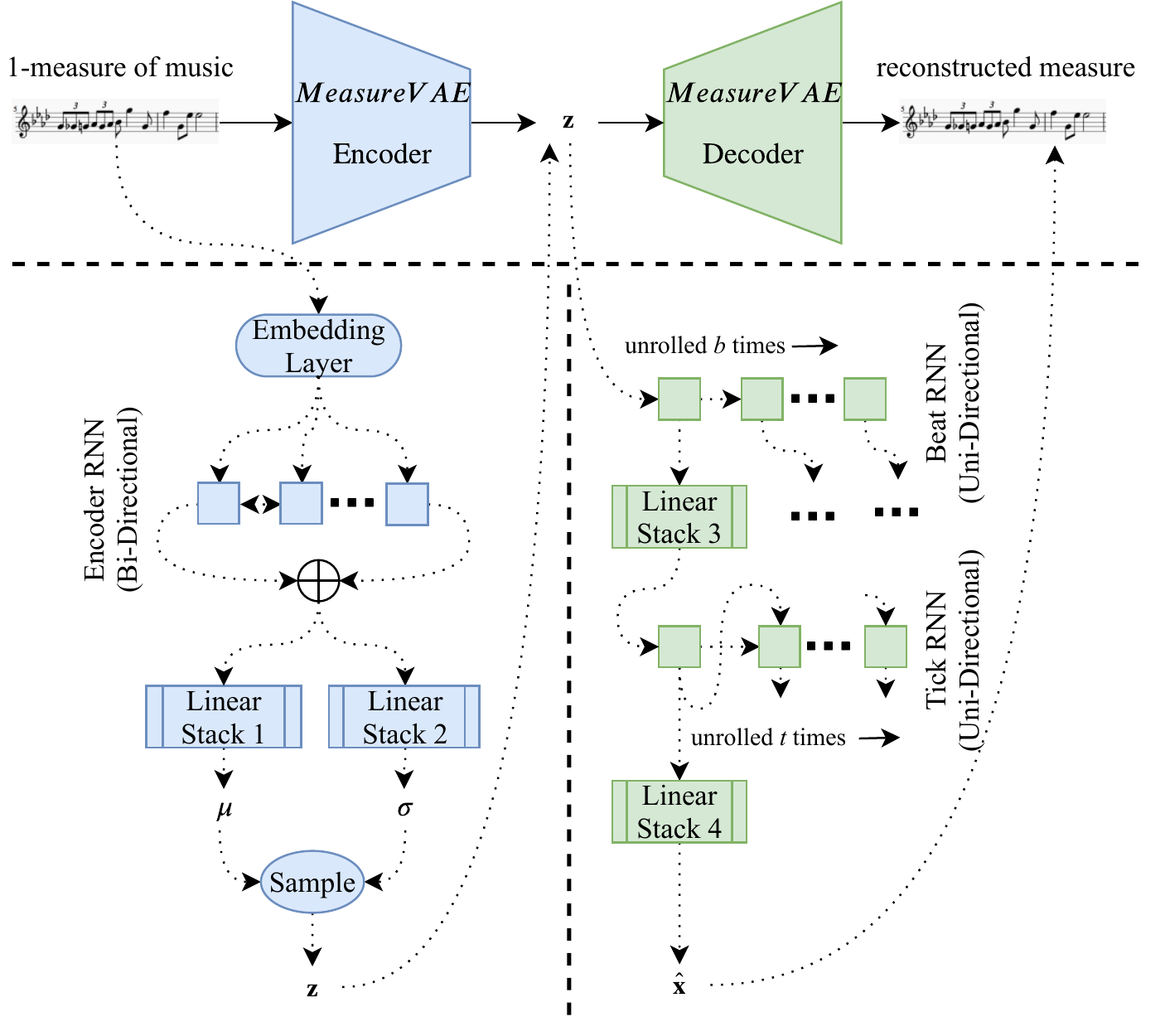}}
        \caption{MeasureVAE schematic. Individual components of the encoder and decoder are shown below the main blocks (dotted arrows indicate data flow within the individual components). $\mathbf{z}$ denotes the latent vector and $\hat{\mathbf{x}}$ denotes the reconstructed measure.}
    \label{fig:measurevae_schematic}
    \end{figure}

\subsection{Model Architectures} \label{modelarch}

\subsubsection{MeasureVAE}
    The MeasureVAE architecture (see \figref{fig:measurevae_schematic}) is loosely based on the hierarchical recurrent MusicVAE architecture~\cite{pmlr-v80-roberts18a} which proved successful in modeling individual measures of music. 
    
    The encoder consists of a learnable embedding layer (operating on tick-level) followed by a bi-directional RNN~\cite{schuster1997bidirectional}. The concatenated hidden state from both directions of the RNN is then passed through two identical parallel linear stacks to obtain the mean $\mu$ and variance $\sigma$ which are used to sample the latent vector $\mathbf{z}$ via $\mathbf{z} \sim \mathcal{N}(\mu, \sigma^{2})$. 
    
    The decoder follows a hierarchical structure where the sampled latent vector $\mathbf{z}$ is used to initialize the hidden state of a beat-RNN which is unrolled $b$ times (where $b$ is the number of beats in a measure). The output at each step of the beat-RNN is passed through a linear stack before being used to initialize the hidden state of a tick-RNN which is unrolled $t$ times (where $t$ is the number of events/ticks in a beat). The outputs of the tick-RNN are individually passed through a second linear stack which maps them back to the data-space. The hierarchical architecture mitigates the auto-regressive nature of the RNN and forces the decoder to use the latent vector more efficiently~(as advocated in \cite{pmlr-v80-roberts18a}). 

    \begin{figure}[t]
        \centerline{
        \includegraphics[width=0.7\columnwidth]{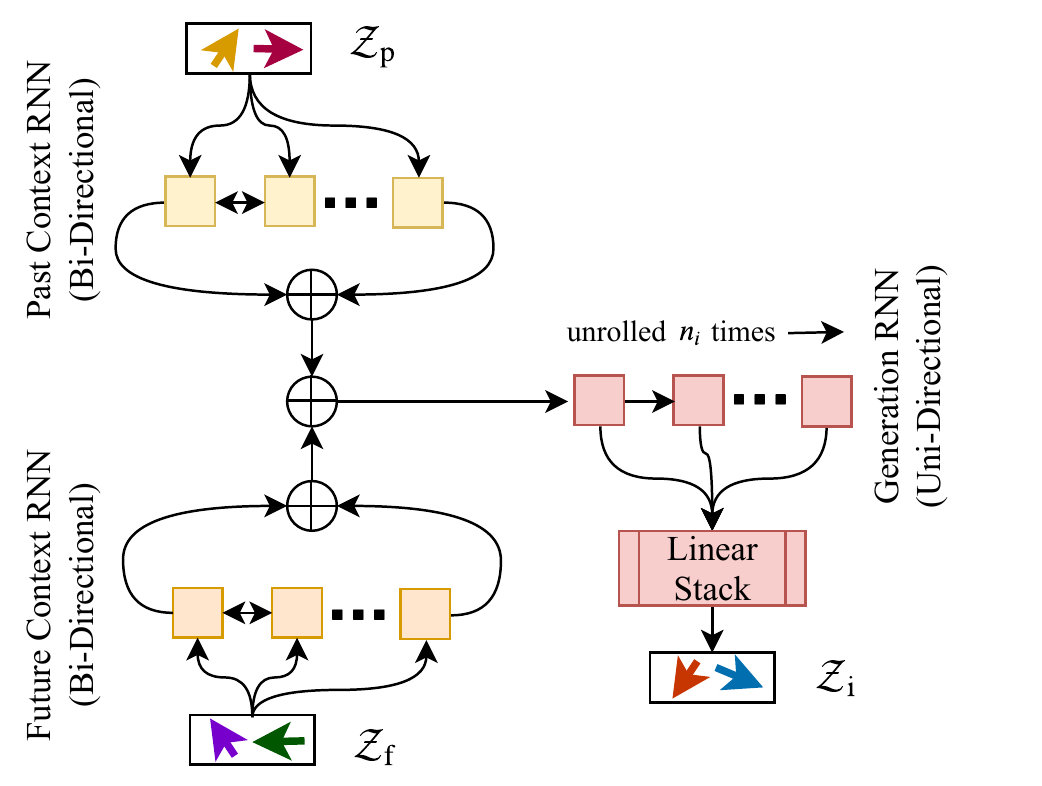}}
        \caption{LatentRNN schematic. The Past-Context and Future-Context-RNNs encode $\mathcal{Z}_\mathrm{p}$ and $\mathcal{Z}_\mathrm{f}$, respectively. The Generation-RNN initialized using a concatenation of context-RNNs embeddings is unrolled $n_\mathrm{i}$ times to get $\mathcal{Z}_\mathrm{i}$.}
    \label{fig:latentrnn_schematic}
    \end{figure}

\subsubsection{LatentRNN}
    The LatentRNN model (see \figref{fig:latentrnn_schematic}) consists of $3$ sub-components. There are $2$ identical bi-directional RNNs, referred to as Past-Context-RNN and Future-Context-RNN, which process the latent vector sequences for the past and future contexts ($\mathcal{Z}_\mathrm{p}$ and $\mathcal{Z}_\mathrm{f}$), respectively. These are unrolled for $n_\mathrm{p}$ and $n_\mathrm{f}$ times in order to encode the context sequences, respectively. The final hidden states of the two context-RNNs are concatenated and then used to initialize the hidden state of a third RNN, referred to as the Generation-RNN, which is unrolled $n_\mathrm{i}$ times. The outputs of the Generation-RNN are passed through a linear stack to obtain $n_\mathrm{i}$ latent vectors corresponding to the inpainted measures. 

    The hyper-parameters for the model configurations are chosen based on initial experiments and are provided in \tabref{tab:model_config}. For the RNN layers in both models, Gated Recurrent Units (GRU)~\cite{jozefowicz2015empirical} are used. 


    \begin{table}[t]
        \footnotesize
    \begin{center}
    \begin{tabularx}{\columnwidth}{Xl}
        \toprule
        \multicolumn{2}{c}{\textit{Measure VAE}} \\ 
        \toprule
        Embedding Layer  & i=dict size, o=10   \\ \midrule
        EncoderRNN       & n=2, i=10, h=512, d=0.5  \\ \midrule
        \begin{tabular}[c]{@{}l@{}}Linear Stack 1 \\ Linear Stack 2\end{tabular} & i=1024, o=256, n=2, non-linearity=SELU \\ \midrule
        BeatRNN          & n=2, i=1, h=512, d=0.5   \\ \midrule
        TickRNN          & n=2, i=522, h=512, d=0.5  \\ \midrule
        Linear Stack 3   & i=512, o=1024, n=1, non-linearity=ReLU \\ \midrule
        Linear Stack 4   & i=512, o=dict size, n=1, non-linearity=ReLU \\ \toprule
        \multicolumn{2}{c}{\textit{Latent RNN}}  \\ 
        \toprule
        \begin{tabular}[c]{@{}l@{}}Past-Context-RNN \\ Future-Context-RNN \end{tabular} & n=2, i=256, h=512, d=0.5  \\ \midrule
        Generation RNN  & n=2, i=1, h=1024, d=0.5  \\ \midrule
        Linear Stack    & i=2048, o=256, n=1, non-linearity=None  \\ \bottomrule
    \end{tabularx}
    \end{center}
    \caption{Table showing configurations of both models. n: Number of Layers, i: Input Size, o: Output Size, h: Hidden Size, d: Dropout Probability, SELU: Scaled Exponential Linear Unit \cite{klambauer2017self}, ReLU: Rectifier Linear Unit}
    \label{tab:model_config}
    \end{table}

\subsection{Stochastic Training Scheme} \label{train_meth}
    We propose a novel stochastic training scheme for training the model. For each training batch, the number of measures to be inpainted $n_\mathrm{i}$ and the number of measures in the past context $n_\mathrm{p}$ are randomly sampled from a uniform distribution. Thus, the number of measures in the future context becomes $n_\mathrm{f} = N - n_\mathrm{i} - n_\mathrm{p}$, where $N$ is the total number of measures in each sequence of the training batch. Using these, the input sequences are split into past, future and target sequences and the model is trained to predict the target sequence given the past and future context sequences. This stochastic training scheme ensures that the model learns to deal with variable length contexts and can perform inpaintings at arbitrary locations. 

\subsection{Data Encoding Scheme} \label{datarep}
    We use a variant of the encoding scheme proposed by Hadjeres et al.~\cite{hadjeres2017deepbach} for our data representation. The original encoding scheme quantizes time uniformly using the sixteenth note as the smallest sub-division. For each sub-division or tick, the note which starts on that tick is represented by a token corresponding to the note name. If no note starts on a tick, a special continuation symbol `\_\_' is used to denote that the previous note is held. Rest is considered as a note and has a special token. 
    The main advantages of this encoding scheme are
    \begin{inparaenum}[(i)]
        \item it uses only a single sequence of tokens, and
        \item uses real note names (e.g., separate tokens for A\# and Bb) which allows generation of readable sheet music.
    \end{inparaenum}
    
\begin{figure}[t]
        \centerline{
        \includegraphics[width=\columnwidth]{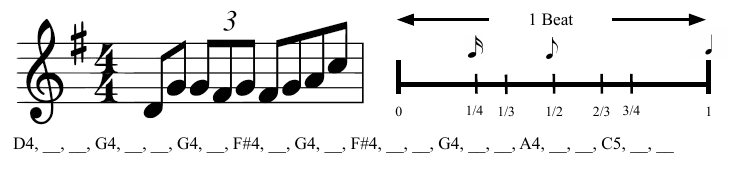}}
        \caption{Figure showing the data representation. The token string on the bottom demonstrates the encoding scheme for the measure displayed on the top-left. Top-right shows the proposed uneven tick-duration scheme for each beat.}
    \label{fig:data_rep}
    \end{figure}

    However, a limitation of using the sixteenth note as the smallest sub-division is that it cannot encode triplets. The naive approach of evenly subdividing the sixteenth note divisions to encode triplets increases the sequence length a factor of $3$ which can make the sequence modeling task harder. To mitigate this limitation, we propose a novel uneven subdivision scheme. Each beat is divided into $6$ uneven ticks (shown in \figref{fig:data_rep}). This allows encoding triplets while only increasing the sequence length by a factor of $1.5$. Consequently, each $4/4$ time signature measure is a sequence of $24$ tokens. 

\section{Experiments} \label{experiments}
    The proposed method is compared with two baseline methods (see \secref{baselines}) using a dataset of monophonic folk melodies in the Scottish and Irish style taken from the Session website~\cite{sturm2016music}. 
    For the purposes of this work, only melodies with $4/4$ time signature in which the shortest note is greater than or equal to the sixteenth note are considered resulting in approx.~$21000$ melodies. 
    Implementation details and source code are available online.\footnote{https://github.com/ashispati/InpaintNet}

\subsection{Baseline} \label{baselines}
    The performance of the proposed method is compared with the AnticipationRNN model proposed by Hadjeres et al.~\cite{Hadjeres2018ARNN}. This model, referred to as \textit{Base-ARNN}, uses a stack of $2$ LSTM-based \cite{hochreiter1997long} RNN layers. Each of the $2$ RNNs comprises of $2$ layers with a hidden size of $256$. In addition to the note-sequence tokens, this model also uses additional metadata information, i.e., tokens to indicate beat and down-beat locations as part of the user-defined constraints. For more details, the readers are directed to~\cite{Hadjeres2018ARNN}.

    The original model operates on tick-level sequences and inpainting locations are specified in terms of individual tick locations. Hence, the inpainting locations may or may not be contiguous. In order to make a fair comparison, a second variant of the AnticipationRNN model is considered, referred to as \textit{Reg-ARNN}, where the stochastic training scheme from \secref{train_meth} is used instead.

\subsection{Training Configuration} \label{train_config}
    The MeasureVAE model was pre-trained using single measures following the standard VAE optimization equation~\cite{kingma2013vae} with the $\beta$-weighting scheme~\cite{higgins2017beta,bowman2015generating}. In order to prioritize high reconstruction accuracy, a low value of $\beta =$ $1$e$-3$ was used. Pre-training was done for $30$ epochs resulting in a reconstruction accuracy of approx.~$99\%$. While this seems to be better than results in \cite{pmlr-v80-roberts18a}, we attribute this to the shorter duration of generation (single measures) and the differences in datasets and data encoding. MeasureVAE parameters were frozen after pre-training and no gradient-based updates were performed on these parameters during the InpaintNet model training.

    The Adam algorithm~\cite{kingma2014adam} was used for model training, with a learning rate of $1$e$-3$, $\beta _1 = 0.9$, $\beta _2 = 0.999$, and $\epsilon=1$e$-8$. To ensure consistency, all models were trained for $100$ epochs (with early-stopping) with the same batch-size using a sub-sequence length of $16$ measures ($384$ ticks). For the InpaintNet and Reg-ARNN models, the number of measures to be inpainted and the number of past measures were randomly selected: $n_\mathrm{i} \in [2,6]$, $n_\mathrm{p} \in [1,16-1-n_\mathrm{i}]$. This ensured that past and future contexts each contain at least $1$ measure. For the baseline models, teacher-forcing was used with a probability of $0.5$. 

\subsection{Predictions on Test Data}
    Two experiments were conducted to evaluate the predictive power of the models. 

    The first experiment considered the average token-wise negative log-likelihood (NLL) on a held-out test set. The results (see first 3 rows of \tabref{tab:nll}) indicate that our proposed model outperforms both baselines, showing an improvement of approx.~$25\%$ in the NLL over the Reg-ARNN model and approx.~$55\%$ over the Base-ARNN model. 

    \begin{table}[t]
        \footnotesize
        \begin{center}
            \begin{tabularx}{0.7\columnwidth}{Xlc}
                \toprule
                \textbf{Model Variant} & \textbf{Test NLL} \\ \toprule
                Base-ARNN & 0.662 \\ \midrule
                Reg-ARNN & 0.402 \\ \midrule
                InpaintNet (Our Method) & \textbf{0.300} \\ \midrule
                PastInpaintNet & 0.643 \\ \midrule
                FutureInpaintNet & 0.481 \\ \bottomrule
            \end{tabularx}
        \end{center}
        \caption{Table showing the average token-wise NLL (nats/token) on the held-out test set (lower is better). InpaintNet outperforms both baselines. The last two rows show the results for the ablation models described in \secref{ablation}.}
    \label{tab:nll}
    \end{table}

\begin{figure}[t]
        \centerline{
        \includegraphics[width=\columnwidth]{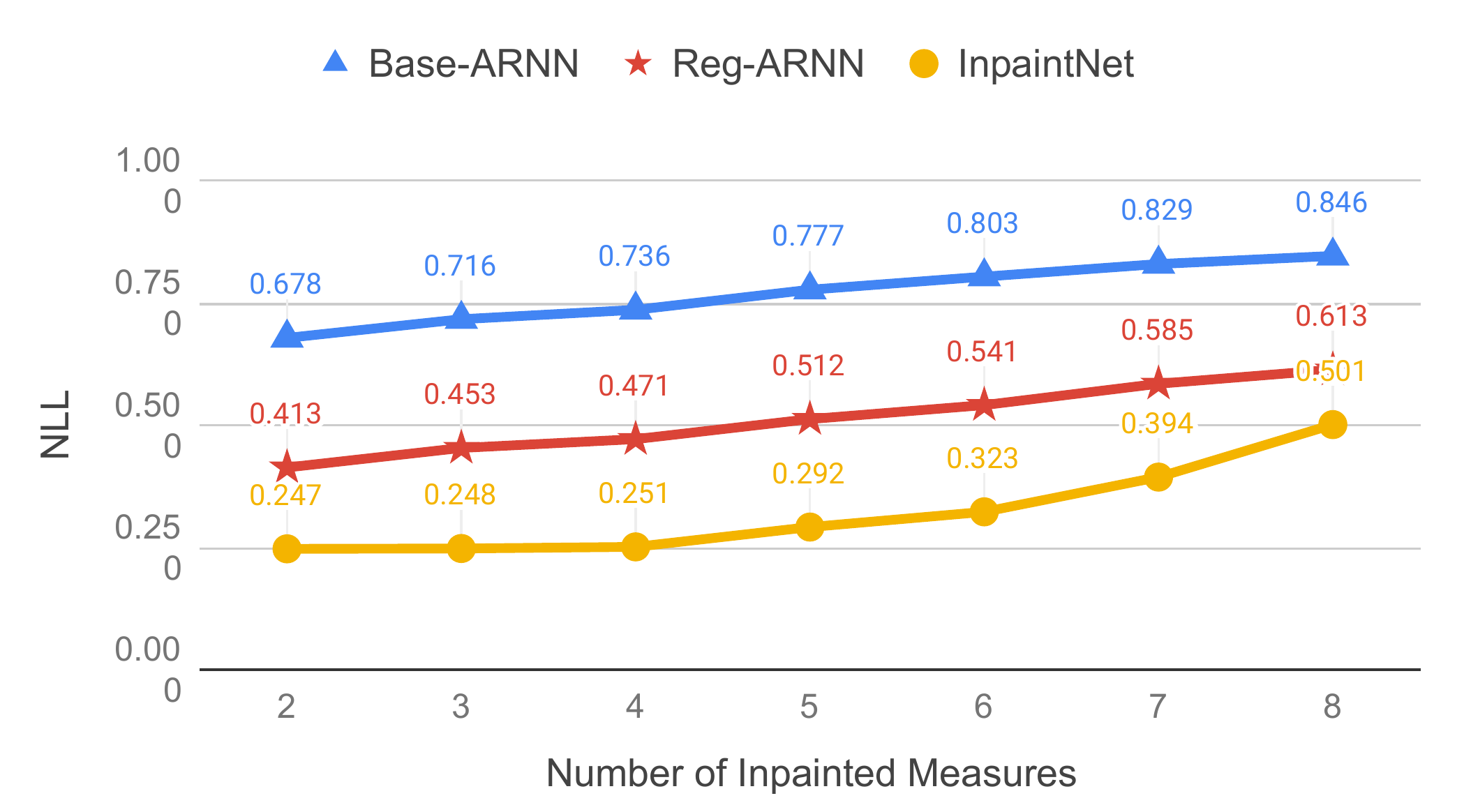}}
        \caption{Figure showing token-wise NLL (nats/token) for different number of inpainted measures on the held-out test set (lower is better). InpaintNet outperforms both baselines. Models were trained to predict only $2$ to $6$ measures. }
    \label{fig:n_bar}
    \end{figure}

    The next experiment compared the models by varying the number of measures to be inpainted. \figref{fig:n_bar} shows the average token-wise NLL when $n_\mathrm{i}$ was increased from $2$ to $8$. Again, our proposed model outperforms both baselines. It should be noted that since the sub-sequence length is constant at $16$ measures, increasing $n_\mathrm{i}$ means that the available context is reduced. Thus, there is an expected drop in the performance with increasing $n_\mathrm{i}$ as the models are forced to make longer predictions with less contextual information. However, the InpaintNet model performs better even when forced to predict beyond the training limit of $6$ measures.

\subsection{Ablations Studies} \label{ablation}
    In order to further ascertain the efficiency of the proposed approach, ablation studies were conducted to evaluate the benefit of adding past and future context information. Specifically, we trained two variants of the InpaintNet model which relied on only one type of contextual information. The first model, referred to as PastInpaintNet only considered the past context $\mathcal{C}_\mathrm{p}$ as input whereas the second model, referred to as FutureInpaintNet considered only the future context $\mathcal{C}_\mathrm{f}$. The last two rows of \tabref{tab:nll} summarize the performance of these ablation models. It is clear that both past and future contexts are important for the modeling process. In addition, we also tried training a variant of the InpaintNet model with an untrained (randomly initialized) MeasureVAE model. This model failed to train properly achieving an NLL of approx.~$1.33$. This indicates that a structured latent space where latent vectors are trained to encode hidden data attributes is important for training the LatentRNN model.


\subsection{Qualitative Analysis} \label{sec:qual_analysis}
    \begin{figure*}[t]
        \centerline{
        \includegraphics[width=\textwidth]{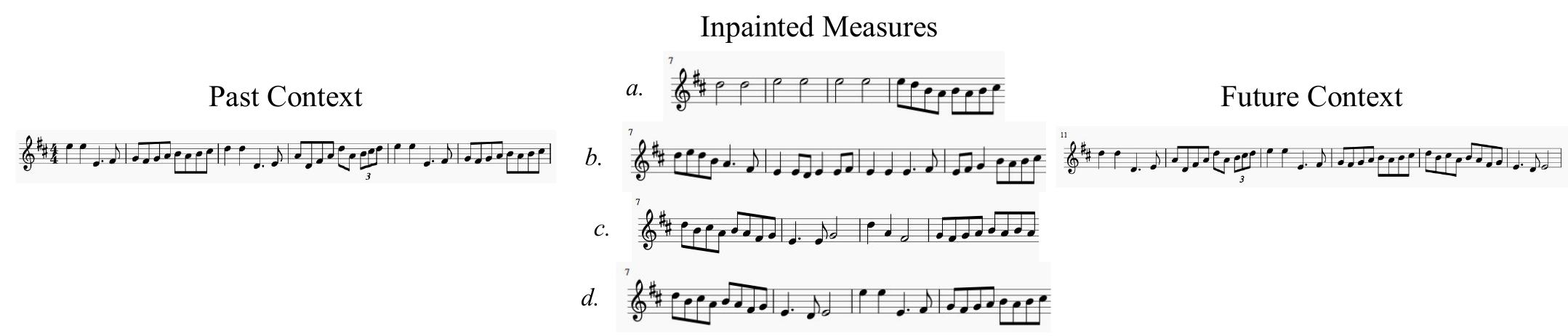}}
        \caption{Figure showing the inpaintings generated by different models for the same context. From top to bottom~---~ \textit{a.}: Base-ARNN, \textit{b.}: Reg-ARNN, \textit{c.}: InpaintNet, \textit{d.}: Original Melody.}
        \label{fig:inpaint_example}
    \end{figure*}
    
    Considering that we are primarily interested in the aesthetic quality of the inpaintings, we encourage the readers to browse through the inpainting examples provided in the supplementary material.\footnote{https://ashispati.github.io/inpaintnet/} We consider some of those examples in the analysis below.

    \figref{fig:inpaint_example} shows sample inpaintings by the models for one of the melodies in the test set. While the Base-ARNN model collapses to produce long half notes which do not effectively reflect the surrounding context, the other two models do better. Both the Reg-ARNN and InpaintNet model generate rhythmically consistent inpaintings. The InpaintNet, in particular, mimics the rhythmic properties of the context better. For instance, measures $7$ and $10$ of the inpainted measures match the rhythm of measures $6$, $14$, and $15$. Also, measure $8$ matches measure $16$. However, the use of G (subdominant scale degree in D-major) in the half-note to end measure $8$ is unusual. We observed that in other examples also, the InpaintNet model occasionally produces pitches which are anomalous~---~either out-of-key or not fitting in the context. The Reg-ARNN model, on the other hand, tends to stay in key. 
    Additional examples are provided in the supplementary material. 
    

    \begin{figure}[t]
        \centerline{
        \includegraphics[width=0.6\columnwidth]{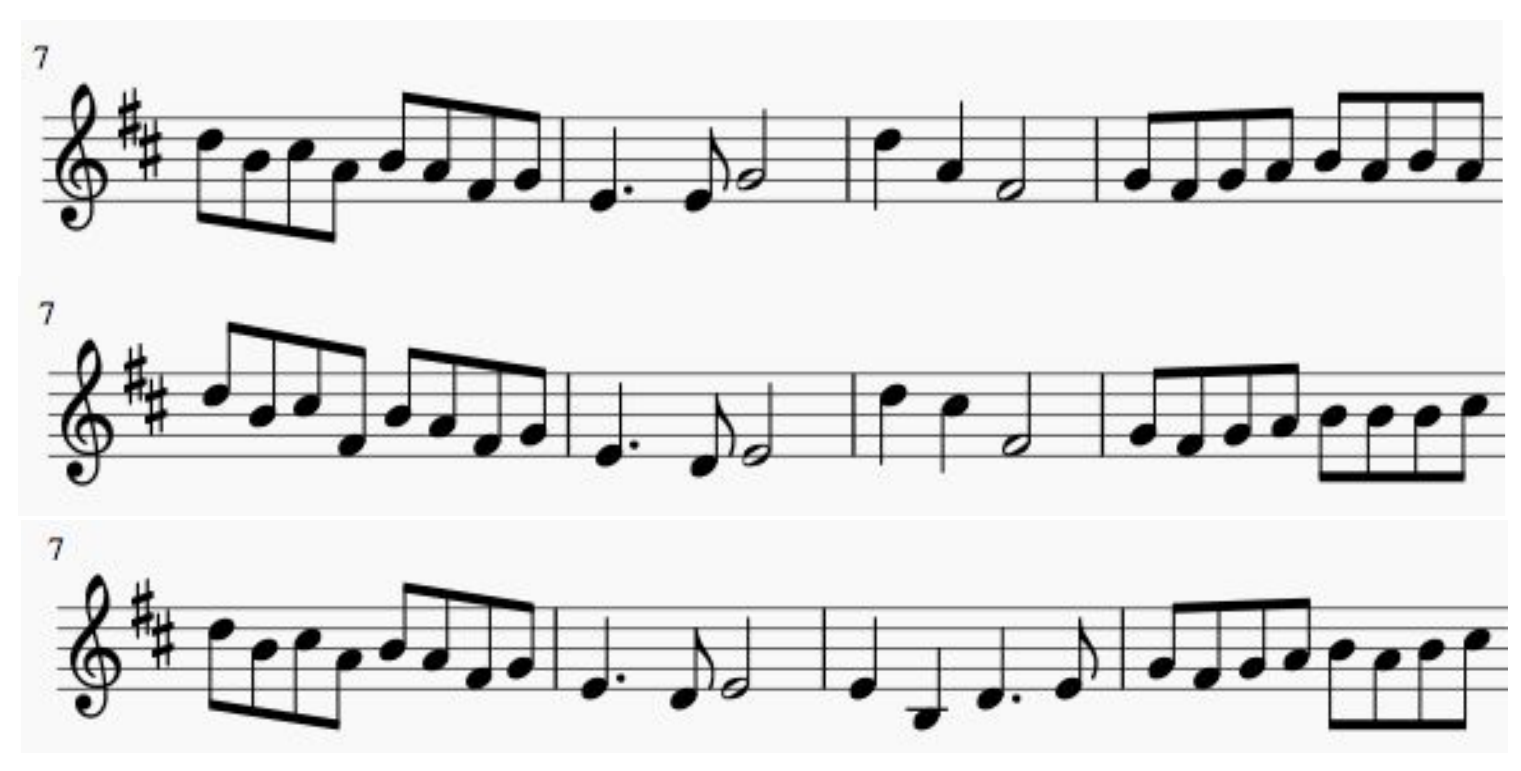}}
        \caption{Figure showing different inpaintings (using the InpaintNet model) for the same context as \figref{fig:inpaint_example}.}
    \label{fig:inpaint_ex2}
    \end{figure}

    One advantage of working with the latent space is that the sampling operation, inherent in the VAE inference process, ensures that for the same context we can get different inpainting results. \figref{fig:inpaint_ex2} shows three such generations for the context of \figref{fig:inpaint_example}. It is interesting to note that the base rhythm is retained across all three inpaintings. This feature is particularly interesting from an interactive music generation perspective, as this model can be used to quickly provide users with multiple ideas and will be investigated further in future work.

    \begin{figure}[t]
        \centerline{
        \includegraphics[width=\columnwidth]{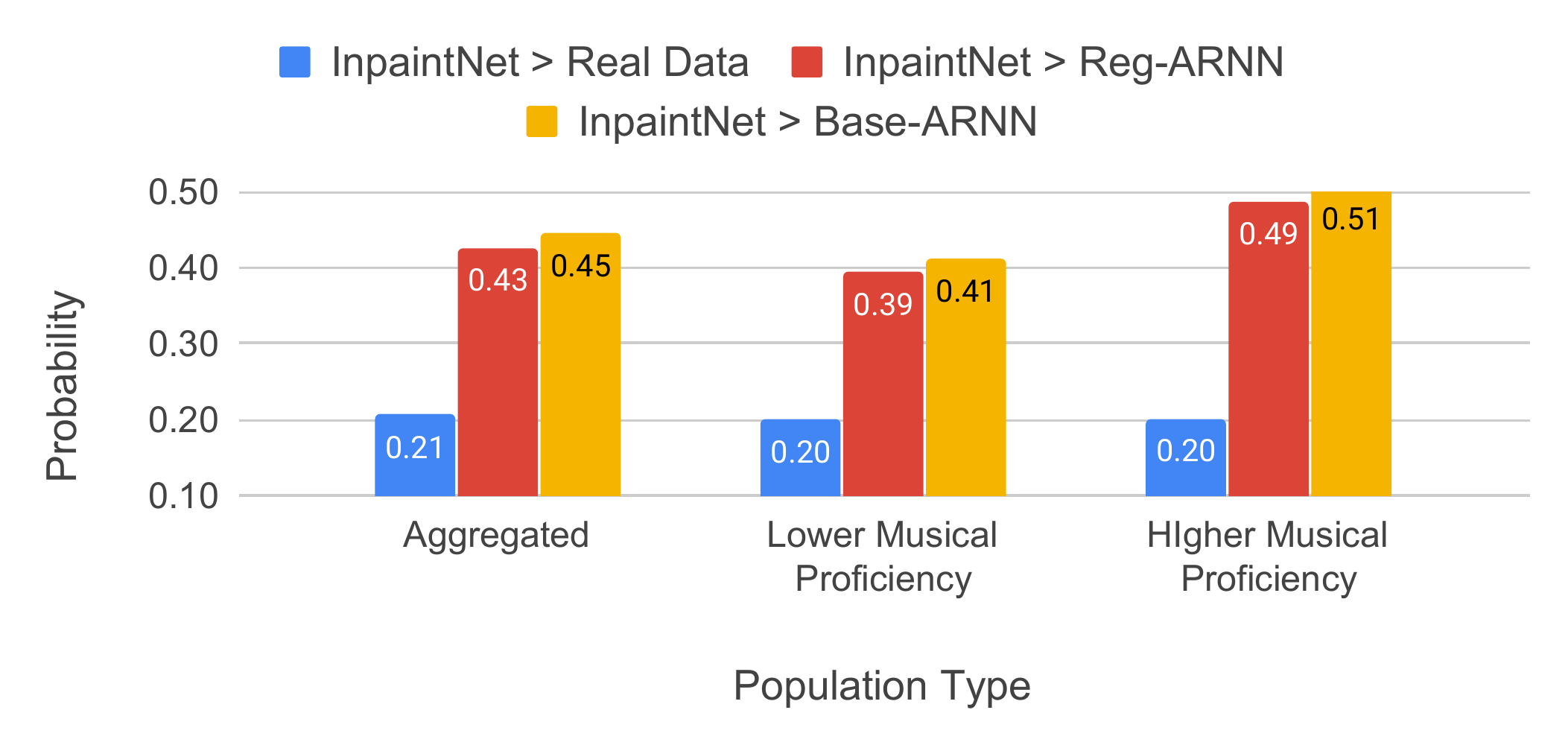}}
        \caption{Figure based on the subjective listening study showing the probability that the InpaintNet model is rated higher. Analysis is based on the Bradley-Terry model \cite{bradley1952rank,hunter2004mm}. The proposed model loses against the real data but performs at par with the baseline models.}
    \label{fig:bt}
    \end{figure}

\subsection{Subjective Listening Study}
    To evaluate the perceived quality of the inpainted measures, a listening test was conducted to compare our proposed model against the two baselines. A set of $30$ melodies from the held-out test set were randomly selected and their first $16$ measures were extracted. The models were then used to inpaint $4$ measures (measure number $7$ to $10$) in these melodic excerpts. Participants were presented with pairs of melodic excerpts and asked to select the one in which they thought the inpainted measures fit better within the surrounding context. In some of the pairs, one melodic excerpt was the real data (without any inpainting). Each participant was presented with $10$ such pairs. A total of $72$ individuals participated in the study ($720$ comparisons). The location of the inpainted measures was kept consistent across all examples so as to prevent confusion among participants and allow them to focus better on the inpainted measures. 

    The Bradley-Terry model~\cite{bradley1952rank,hunter2004mm} for paired comparisons was used to get an estimate of how the proposed model performs against the baselines and the real data (see \figref{fig:bt}). While the proposed model expectedly has a very low probability of winning against the real data (wins approx.~$1$ out of $5$ times), it performs only at par with the baseline models (with probability approx.~$0.5$). Significance tests using the Wilcoxon signed rank test were further conducted which validated that differences between the proposed model and the baselines were not statistically significant ($p$-value $>0.01$). This was unexpected since the proposed model showed significant improvement over the baselines in the NLL metric. Further dividing the study population into two groups differing in musical proficiency (based on the Ollen index~\cite{ollen2006criterion}) showed that, comparatively, the group with greater musical proficiency favored the generations from the InpaintNet model more than the group with less musical proficiency. 

    Additional analysis revealed that cases where the InpaintNet model performed the worst (maximum losses against the baselines), had anomalies in the predicted pitch similar to those discussed in \secref{sec:qual_analysis}. Specifically, they either had a single out-of-key note (e.g., F note in G-Major scale) or used a pitch or interval not used in the provided contexts. We conjecture that it is these anomalous pitch predictions which lead to poor perceptual ratings in spite of the model performing better in terms of modeling rhythmic features. This will be analyzed further in future studies.

\section{Conclusion} \label{conclusion}
    This paper investigates the problem of musical score inpainting and proposes a novel approach to generate multiple measures of music to connect two musical excerpts by using a conditional RNN which learns to traverse the latent space of a VAE. We also improve upon the data encoding and introduce a stochastic training process which facilitate model training and improve generalization. 
    The proposed model shows good performance across different objective and subjective evaluation experiments. The architecture also enables multiple generations with the same contexts, thereby, making it suitable for interactive applications \cite{hadjeres18agnostic}. 
    We think the idea of learning to traverse latent spaces could be useful for other music generation tasks also. For instance, the architecture of the LatentRNN model can be changed to add contextual information from other voices/instruments to perform multi-instrument music generation. Future work will include a more thorough investigation of the anomalies in pitch prediction. A possible way to address that would be to add the context embedding as input at each step of unrolling the LatentRNN or use additional regularizers. Another promising avenue for future work is substituting RNNs with attention-based models~\cite{vaswani2017attention} which have had success in sequential music generation tasks \cite{huang2019musictransformer}.
    
\newpage
\bibliography{Inpainting}

\end{document}